\documentclass[runningheads]{llncs}
\usepackage[T1]{fontenc}
\usepackage{graphicx}
\usepackage{multirow}
\usepackage[outdir=./]{epstopdf}
\usepackage{subcaption}
\usepackage{caption} 
\usepackage{amsfonts}
\usepackage{amsmath}
\usepackage{booktabs}
\usepackage[table,xcdraw]{xcolor}
\begin{document}
%
%\title{Contribution Title\thanks{Supported by organization x.}}
\title{HistoSmith: Single-Stage Histology Image-Label Generation via Conditional Latent Diffusion for Enhanced Cell Segmentation and Classification}

%\name{Valentina Vadori$^{1}$ \ Jean-Marie Graïc$^{2}$ \ Livio Finos$^{3}$ \ Livio Corain$^{4}$ \ Antonella Peruffo$^2$ \ Enrico Grisan$^{1}$ }
%\address{$^{1}$London South Bank University, School of Engineering, United Kingdom
%\\ $^{2}$University of Padova, Dept. of Comparative Biomedicine \& Food Science, Italy
%\\ $^{3}$University of Padova, Dept. of Developmental Psychology and Socialisation, Italy
%\\ $^{4}$University of Padova, Dept. of Management and Engineering, Italy}

%
\titlerunning{HistoSmith: Single-Stage Histology Image-Label Generation}
%\author{Valentina Vadori\inst{1}\orcidID{0009-0004-8043-741X} \and
%Antonella Peruffo\inst{2}\orcidID{0000-0003-0454-9087} \and
%Jean-Marie Graïc\inst{2}\orcidID{0000-0002-1974-8356} \and
%Livio Finos\inst{3}\orcidID{0000-0003-3181-8078} \and
%Livio Corain\inst{4}\orcidID{0000-0002-8104-1255} \and
%Enrico Grisan\inst{1}\orcidID{0000-0002-7365-5652}}

\author{Valentina Vadori\inst{1} \and
Jean-Marie Graïc\inst{2} \and
Antonella Peruffo\inst{2} \and
%Jean-Marie Graïc\inst{2} \and
%Giulia Vadori\inst{2} \and
Livio Finos\inst{3} \and
Ujwala Kiran Chaudhari\inst{1} \and
Enrico Grisan\inst{1}}

\authorrunning{Vadori et al.}
\institute{Dept. of Computer Science \& Informatics, London South Bank University, UK\\
\email{\{vvadori,egrisan\}@lsbu.ac.uk} \and
Dept. of Comparative Biomedicine \& Food Science, University of Padova, IT\and
Dept. of Statistical Sciences, University of Padova, IT}
%\institute{Dept. of Computer Science \& Informatics, London South Bank University, UK\\
%\email{\{vvadori,egrisan\}@lsbu.ac.uk} \and
%Dept. of Comparative Biomedicine \& Food Science, University of Padova, IT\\
%\email{\{antonella.peruffo,jeanmarie.graic,giulia.vadori\}@unipd.it} \and
%Dept. of Statistical Sciences, University of Padova, IT \\
%\email{livio.finos@unipd.it}}

%\institute{London South Bank University, School of Engineering, United Kingdom  \and
%University of Padova, Dept. of Comparative Biomedicine \& Food Science, Italy

%
\maketitle              % typeset the header of the contribution
\begin{abstract}
Precise segmentation and classification of cell instances are vital for analyzing the tissue microenvironment in histology images, supporting medical diagnosis, prognosis, treatment planning, and studies of brain cytoarchitecture. However, the creation of high-quality annotated datasets for training remains a major challenge. 
This study introduces a novel single-stage approach (\textit{HistoSmith}) for generating image-label pairs to augment histology datasets. Unlike state-of-the-art methods that utilize diffusion models with separate components for label and image generation, our approach employs a latent diffusion model to learn the joint distribution of cellular layouts, classification masks, and histology images. This model enables tailored data generation by conditioning on user-defined parameters such as cell types, quantities, and tissue types.
Trained on the Conic H\&E histopathology dataset and the Nissl-stained CytoDArk0 dataset, the model generates realistic and diverse labeled samples. Experimental results demonstrate improvements in cell instance segmentation and classification, particularly for underrepresented cell types like neutrophils in the Conic dataset. %(<1\% of cells). 
These findings underscore the potential of our approach to address data scarcity challenges.  %in histology analysis.

\keywords{Cell Segmentation \and Cell Classification \and Histology \and Data Augmentation \and Conditional Generative Models \and Latent Diffusion}
\end{abstract}
\section{Introduction}
Deep learning (DL) can be applied directly to histological sections for diagnostic or prognostic purposes, but this approach often suffers from limited interpretability. Alternatively, using DL to detect, segment, and classify cells or nuclei allows for the extraction of interpretable biomarkers that characterize the tissue microenvironment through cell-related features such as quantity, size, morphology, and spatial arrangement. These biomarkers, in turn, enable downstream applications including disease detection, outcome prediction, and the design of personalized treatments \cite{song2023artificial,nunes2024survey}. In neuroscience, the segmentation and classification of neurons and glial cells in whole-slide images of brain tissue facilitate the precise quantification of \textit{brain cytoarchitecture}\footnote{The cellular organization of brain cells, including their types, densities, and spatial arrangements.}. Cytoarchitectural analysis are essential for advancing our understanding of the pathogenesis of neurodegenerative and neuroinflammatory disorders, while also providing insights into brain development and evolution \cite{AMUNTS20071061,falcone2021neuronal,graic2023cytoarchitectureal,meystre2024cell,graic2024age}. DL methods for cell instance segmentation (CS) and cell classification (CC) \footnote{CS: identifying and outlining individual cells; CC: recognizing cell types} rely heavily on supervised learning \cite{song2023artificial}, which requires large labeled datasets \cite{schmidt2018cell,graham2019hover,stringer2021cellpose,vadori2024cisca}. Yet, labeling cell boundaries and cell types is labor-intensive and demands expert input, limiting the availability of annotated data. Generative models address this challenge by generating synthetic images with segmentation masks, yielding additional data \cite{chen2024comprehensive}.%,falcone2021neuronal,graic2023cytoarchitectureal,meystre2024cell,graic2024age}.

\cite{tasnadi2023structure} employ two generative adversarial networks (GANs): the first generates heat-flow encodings, which are post-processed into labeled masks, where each cell is assigned a unique integer, and the second performs image-to-image translation to create corresponding microscopy images. Although GANs can produce high-quality images, their adversarial training process often results in unstable training dynamics and restricted diversity in the generated outputs \cite{dhariwal2021diffusion}. Diffusion models, especially the denoising diffusion probabilistic model (DDPM) \cite{ho2020denoising}, have recently gained prominence as a superior alternative. 
%However, their application to this specific task remains relatively unexplored.
%to GANs. 
DDPMs are likelihood-based models that progressively add noise to an image in the \textit{diffusion} process and learn to reverse this process. At inference, a random noise input is gradually  mapped onto a realistic output \cite{luo2022understanding}. \cite{yu2023diffusion} use two DDPMs: the first generates nuclei layouts as binary masks and distance maps and the second synthesizes histopathology images conditioned on them. The synthetic samples are added to the real dataset for training a CS model. It is shown that carefully selecting 10\% of the labeled real dataset and augmenting it with synthetic samples can achieve CS performance comparable to using the  real dataset. Augmenting the full real dataset results in improvements in Dice scores ranging from 0.6\% to 1.3\%. \cite{min2025co} apply a DDPM to generate histopathology images, distance maps, and semantic masks for CS and CC. The DDPM is conditioned on nuclei centroid layouts and structure-related textual prompts, such as “[tissue type] with [list of nuclei types]”. The layouts appear to be sourced from the training set. In the satellite imaging domain, \cite{toker2024satsynth} train an unconditional DDPM to generate satellite images along with their corresponding semantic masks for semantic segmentation tasks. When evaluated on an Earth observation benchmark dataset augmented with a synthetic dataset of the same size as the real dataset, the mean IoU across five different models shows an average improvement of $2.67\%$.
%The study highlights that generating image-mask pairs at a resolution of $128 \times 128$ and  upscaling them to $256 \times 256$ using a second DDPM conditioned on the lower-resolution outputs leads to improved performance. When applied to an Earth observation benchmark dataset and training semantic segmentation models on the combined dataset $\mathcal{D} \cup \mathcal{D}'$, where $\mathcal{D}'$ is the generated dataset with the same size of the real one, the mean IoU across five different models improves by an average of $2.67\%$. %, with individual gains ranging from $0.67\%$ to $7.59\%$, depending on the model.
%of the origianl earth observation benchmark
\begin{figure}
\centering
\includegraphics[width=0.90\textwidth]{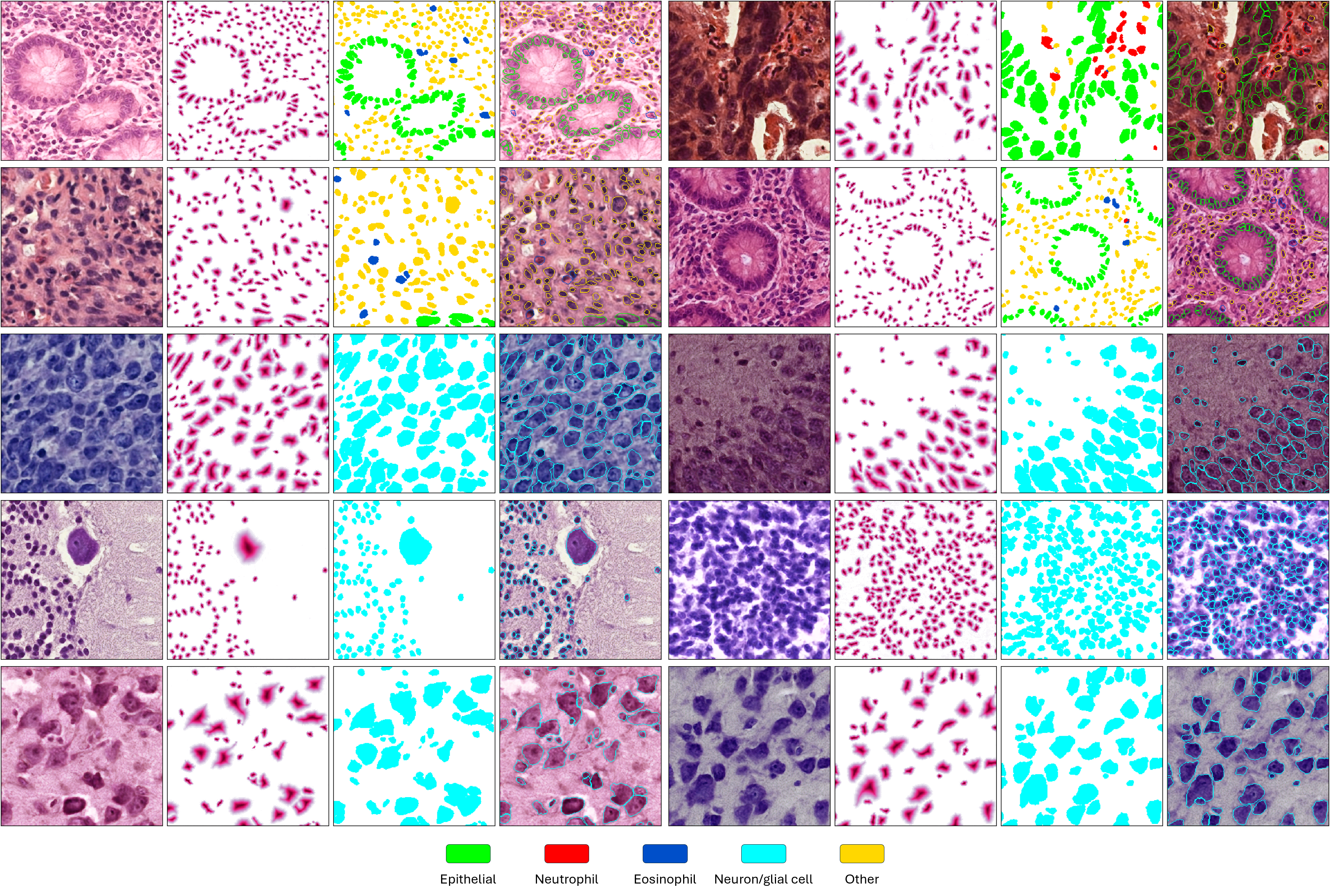}
\vspace{-5pt}
\caption{Ten HistoSmith-generated samples of the colon (rows 1-2), hippocampus (row 3), cerebellum (row 4), and auditory cortex (row 5), each with a generated image, distance map, cell type semantic mask, and post-processed label overlay.} \label{generatedsamples}
\end{figure}
DDPMs operate directly on raw image pixels.
%, which can be computationally intensive. 
Latent diffusion models (LDMs) \cite{rombach2022high}
%address this limitation by 
encode images into a compressed latent representation using a variational autoencoder or similar architecture and perform diffusion operations in this lower-dimensional latent space, reducing computational costs while minimizing noise and redundancy inherent in pixel-level representations.
%. By working in the latent space, LDMs capture the essential features of the data while minimizing noise and redundancy inherent in pixel-level representations. 
The final denoised output is decoded back into the original image space. 

In this study, we present \textit{HistoSmith}, a LDM designed to generate image-label pairs for augmenting CS and CC datasets. Our key contributions include:
(1) A novel LDM for histology data augmentation, capable of generating histology images and their corresponding CS and CC labels simultaneously.
(2) A parameter-driven generative process, conditioned on user-defined parameters such as cell types and quantity, without requiring cell layouts as additional input
%(3) we evaluate HistoSmith's generative capability, particularly under conditions that deviate from those encountered during training, and 
%(e.g., generating images specified to contain only neutrophils).
(3) Evidence demonstrating its effectiveness, showing that augmenting CoNIC and CytoDArk0 datasets with HistoSmith leads to average metric gains of 1.9\% and 3.4\%, respectively, in CS and CC.
(4) A comprehensive analysis of the quality of generated samples.
%(3) A unified, one-step generative approach trained jointly across datasets with different staining methods, as opposed to a two-step approach, where segmentation labels are generated first and then used to condition image generation in a subsequent step.
To the best of our knowledge, HistoSmith is the first LDM-based approach for histology dataset augmentation in CS and CC. Designed to address data scarcity, it can also be used to tackle class imbalance by steering the generative process toward underrepresented cell types, such as inflammatory cells, which are difficult to identify in datasets like CoNIC \cite{graham2024conic,vadori2024cisca}.

\section{Method}
\begin{figure}
\centering
\includegraphics[width=0.9\textwidth]{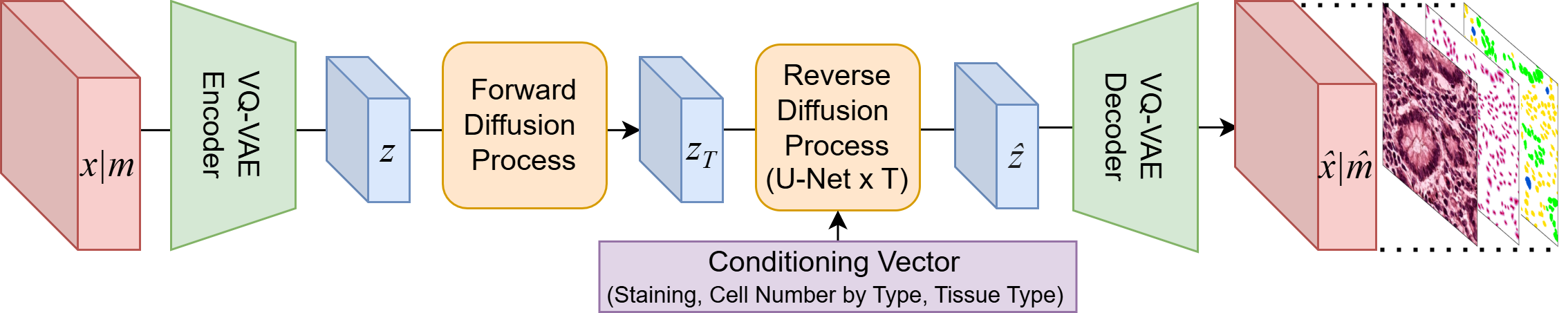}
\vspace{-2pt}
\caption{The proposed HistoSmith framework for generative data augmentation.} \label{HistoSmith}
\end{figure}

Our goal is to augment a labeled image dataset with generated samples to improve CS and CC performance. We propose an LDM \cite{rombach2022high} that concurrently generates a histological image and its masks. As shown in Fig. \ref{generatedsamples}, these include a cellular layout encoded as a distance map and a semantic mask for cell types, post-processed into a label map and unique cell-type assignments via thresholding, morphological operations, and majority voting.
%As shown in Fig. \ref{generatedsamples}, these include a cellular layout encoded as a distance map and a semantic mask, post-processed into a label map and unique cell assignments via thresholding, morphology, and majority voting.
%Our objective is to enhance a dataset with a limited number of labeled images by generating additional samples to improve CS and CC performance. To boost the diversity of labeled images, it is advantageous to synthesize both the images and their corresponding instance maps. We propose a two-step strategy for generating new labeled images, with both steps relying on diffusion models. An overview of the proposed framework is presented in Fig. 2. In this section, we provide a detailed explanation of the two steps.

%As per \cite{rombach2022high} and Fig. \ref{HistoSmith}, a VQ-VAE learns a discrete latent representation \({z}\) from images \({x}\) and masks \({m}\), followed by diffusion model training in this space.
As per \cite{rombach2022high} and Fig. \ref{HistoSmith},  a VQ-VAE is trained on a dataset \(\mathcal{D} = \{({x}_i, {m}_i)\}_{i=1}^{N}\) to learn a discrete latent representation \({z}\) of the input images \({x}\) and corresponding masks \({m}\). A diffusion model is then trained in the learned latent space. 

The VQ-VAE encoder, parameterized by a convolutional neural network (CNN), maps an input image concatenated with its masks along the channel dimension \({x|m}\) to a continuous latent representation, reducing the resolution by a factor of 4. The latent vectors forming this representation are mapped to the nearest prototype in a codebook, yielding a quantized representation \({z}\). A CNN-based decoder reconstructs the original image and masks. Our decoder has two heads: one predicts the input image and distance map, the other the semantic mask of cell types. The training objective for the first head includes a reconstruction and perceptual loss for fidelity, a commitment loss for stable encoder representations, a codebook loss for optimizing discrete representation learning, and an adversarial loss to mitigate blurriness, consistent with previous work. For the second head, inspired by semantic segmentation tasks, we apply a categorical cross-entropy loss and a Tversky loss \cite{salehi2017tversky}.
%Mathematically, the loss function can be expressed as:
%\[
%L_{VQ-VAE} = \|x - \hat{x}\|_2^2 + \beta \|\text{sg}[z_e(x)] - z_q(x)\|_2^2 + \alpha \|z_e(x) - \text{sg}[z_q(x)]\|_2^2,
%\]
%where $\text{sg}[\cdot]$ denotes the stop-gradient operator, and $\alpha, \beta$ are hyperparameters.

In the diffusion model, the forward process gradually adds Gaussian noise to the latent variables \({z}\) over $T$ timesteps with a linear schedule. In the reverse process,  a time-conditional U-Net learns to denoise the latent representations by predicting the added noise at each step. %using
%\[
%\(\mathbf{z}\)_t = \sqrt{\bar{\alpha}_t} \(\mathbf{z}\) + \sqrt{1 - \bar{\alpha}_t} \epsilon, \quad \epsilon \sim \mathcal{N}(0, I),
%\]
%where $\bar{\alpha}_t$ is a predefined noise schedule.
%\[
%\epsilon_\theta(\mathbf{z}_t, t) \approx \epsilon.
%\]
%The U-Net is trained using a variational lower bound objective:
%$L_{LDM} := \mathbb{E}_{{z}, \epsilon \sim \mathcal{N}(0,1), t} \left[ \| \epsilon - \epsilon_{\theta}(z_t, t) \|_2^2 \right]$ as the loss function, with $t$ uniformly sampled from ${1, . . . , T}$. 
For more details we refer to \cite{van2017neural,rombach2022high}.

We condition the reverse diffusion process with an 10-dimensional vector encoding staining type, tissue type, and cellular composition. The first element specifies the staining method (0 for HE, 1 for Nissl). The next five represent normalized cell counts for neutrophils, epithelial cells, eosinophils, other cells, and neuron/glia. The final four use one-hot encoding for colon, auditory cortex, cerebellum, and hippocampus. Cell counts are normalized by the maximum value observed in a training patch to enable global quantity learning. The conditioning vector is mapped to an embedding vector, initialized from a normal distribution, which is then combined with the time embedding to condition the U-Net on the timestep. At inference, a random Gaussian noise sample and the conditioning vector are fed into the reverse diffusion process. After denoising, the VQ-VAE decoder reconstructs a new image and its corresponding masks.

%𝑇
%T timesteps, refining the latent representation. Once denoised, the VQ-VAE decoder reconstructs the final image and masks.

\section{Experiments}
\label{sec:experiments}
\begin{figure}
\centering
\includegraphics[width=0.99\textwidth]{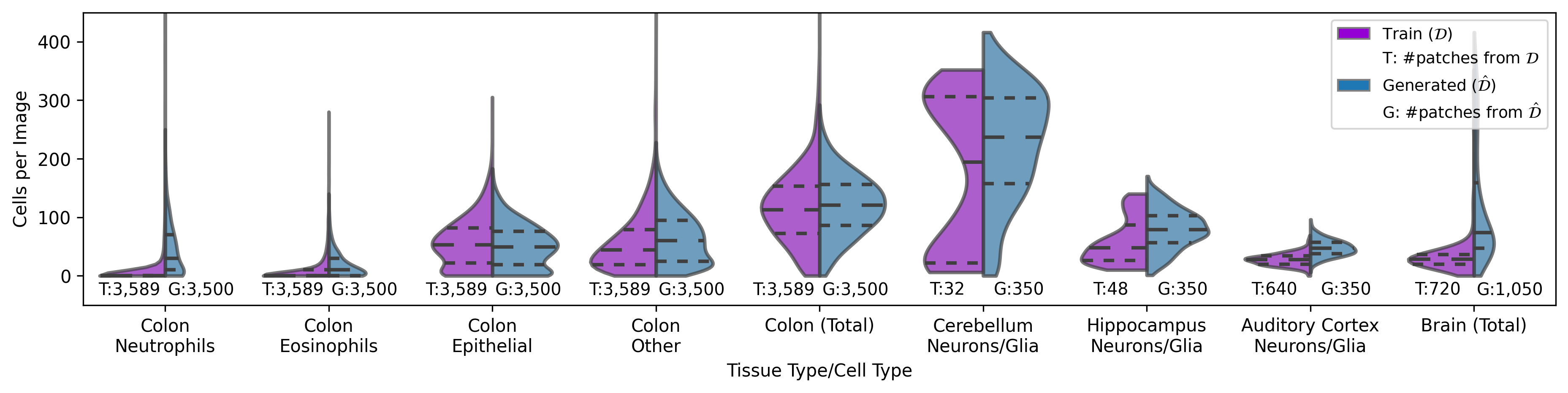}
\vspace{-10pt}
\caption{Cell distribution per image across tissue/cell types in datasets $\mathcal{D}$ and $\hat{\mathcal{D}}$.} \label{cell_counts}
\end{figure}

\begin{table}[]
\caption{CS and CC results on CoNIC and CytoDArk.}
\label{results:overall}
\vspace{-7pt}
\centering
\setlength{\tabcolsep}{1.7pt}
\renewcommand{\arraystretch}{0.9} 
\begin{tabular}{@{}lcccccccc|ccccccc@{}}
\toprule
 & \multicolumn{8}{c|}{\textbf{CoNIC}} & \multicolumn{7}{c}{\textbf{CytoDArk0}} \\ \midrule
\textbf{} & \textbf{Dice} & \textbf{P} & \textbf{R} & \textbf{DQ} & \textbf{SQ} & \textbf{PQ} & \textbf{R2} & \textbf{mPQ$^+$} & \textbf{Dice} & \textbf{P} & \textbf{R} & \textbf{DQ} & \textbf{SQ} & \textbf{PQ} & \textbf{R$^2$} \\
$\mathcal{D}$ & 82.1 & 80.8 & 80.4 & 80.3 & 79.9 & 64.4 & 74.4 & 47.2 & 86.5 & 77.3 & 78.4 & 77.1 & 86.3 & 66.7 & 99.2 \\
$\mathcal{D} \cup \hat{\mathcal{D}}$ & 82.7 & 81.6 & 80.2 & 80.7 & 80.6 & 65.3 & 83.7 & 49.9 & 89.2 & 81.7 & 84.1 & 82.2 & 87.3 & 71.8 & 99.3 \\ \bottomrule
\end{tabular}
\setlength{\tabcolsep}{4pt}
\begin{tabular}{@{}lcccccccccccc@{}}
\textbf{CoNIC} & \multicolumn{3}{c}{\textbf{Neutrophils}} & \multicolumn{3}{c}{\textbf{Epithelial}} & \multicolumn{3}{c}{\textbf{Eosinophils}} & \multicolumn{3}{c}{\textbf{Other}} \\ 
\addlinespace[-3pt]
\midrule
 & \textbf{DQ} & \textbf{SQ} & \textbf{PQ} & \textbf{DQ} & \textbf{SQ} & \textbf{PQ} & \textbf{DQ} & \textbf{SQ} & \textbf{PQ} & \textbf{DQ} & \textbf{SQ} & \textbf{PQ} \\
$\mathcal{D}$ & 29.2 & 75.5 & 22.0 & 77.6 & 77.8 & 60.4 & 50.0 & 77.0 & 38.5 & 81.3 & 83.4 & 67.8 \\
$\mathcal{D} \cup \hat{\mathcal{D}}$ & 38.8 & 75.0 & 29.1 & 77.9 & 78.2 & 60.9 & 52.2 & 77.3 & 40.3 & 82.0 & 84.5 & 69.2 \\ \bottomrule
\textbf{CytoDArk0} & \multicolumn{3}{c}{\textbf{Cerebellum}} & \multicolumn{3}{c}{\textbf{Hippocampus}} & \multicolumn{3}{c}{\textbf{Aud. Cortex}} & \multicolumn{3}{c}{\textbf{Vis. Cortex}} \\ 
\addlinespace[-3pt]
\midrule
$\mathcal{D}$ & 86.4 & 84.7 & 73.2 & 81.5 & 89.3 & 72.8 & 80.8 & 84.4 & 68.3 & 79.0 & 88.1 & 69.6 \\
$\mathcal{D} \cup \hat{\mathcal{D}}$ & 88.5 & 85.7 & 75.8 & 85.8 & 89.5 & 76.8 & 81.9 & 84.6 & 69.4 & 81.3 & 90.0 & 73.2 \\ \bottomrule
\end{tabular}
\end{table}
We assessed HistoSmith’s output quality by training it on a merged dataset comprising the \textit{CoNIC} \cite{graham2024conic} and \textit{CytoDArk0} \cite{vadori2024cytodark0} datasets. CoNIC includes 4,981 H\&E-stained colon tissue histology images with six cell types: neutrophils, epithelial cells, lymphocytes, plasma cells, eosinophils, and connective tissue cells. To simplify differentiation, annotations were modified by grouping lymphocytes, plasma, and other cell types into a single \textit{other} category. CytoDArk0 is the first open dataset of Nissl-stained mammalian brain histology with annotated neurons and glial cells. This study uses the CytoDArk0\_20x\_256 subset, with 1,104 patches. Both datasets contain 256×256 patches at 20× magnification.
We trained HistoSmith using the Adam optimizer for 150 epochs with learning rates of \(1 \times 10^{-5}\) for the VQ-VAE and \(5 \times 10^{-6}\) for the U-Net. We implemented the data splitting, oversampling, and staining augmentation techniques described in \cite{vadori2024cisca}. The source code will be available at \url{https://github.com/vadori/cytoark}.
\begin{figure}
\centering
\includegraphics[width=0.99\textwidth]{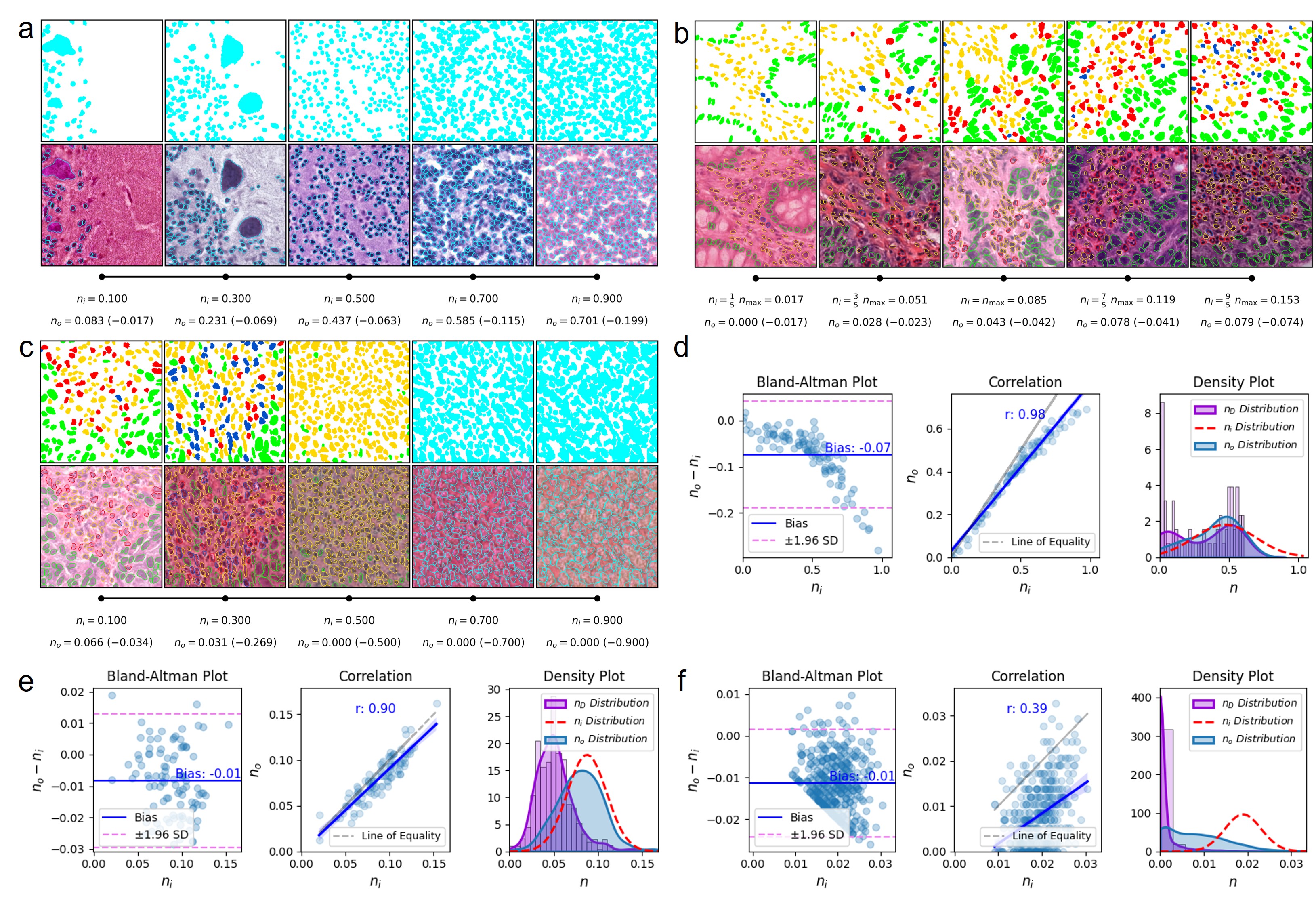}
\vspace{-7pt}
\caption{Correlation between conditioning ($n_i$) and generated cell quantities ($n_o$).} \label{quantityAnalysis}
\end{figure}

After training, a new dataset \(\hat{\mathcal{D}} = \{(\hat{\mathbf{x}}_i, \hat{\mathbf{m}}_i)\}_{i=1}^{M}\) was generated to augment \(\mathcal{D}\) for training the CISCA model \cite{vadori2024cisca} for CS and CC. 
%CISCA has demonstrated superior robustness and accuracy across various staining techniques, magnifications, and tissue types. 
We added 350 brain tissue samples and 3,500 colon samples, approximately doubling the training set size.
For each tissue, cell quantities for conditioning vectors were sampled from Gaussian distributions, with standard deviations matching the training set and means scaled by a multiplicative factor \(\alpha \in [1,2]\) for all cell types and \(\alpha = 20\) for neutrophils and eosinophils. This resulted in an increase in the average cell count per patch in \(\hat{\mathcal{D}}\), as shown in Fig. \ref{cell_counts}, where distributions in \(\hat{\mathcal{D}}\) tend to shift upwards or exhibit longer tails. This provides more training cell instances and enhances the representation of neutrophils and eosinophils, minority classes in the CoNIC dataset. After training, CISCA was evaluated on the test sets of CoNIC and CytoDArk0.

Additionally, 100 extra patches were generated per brain tissue type and 400 for the colon, storing the conditioning vector values to analyze the relationship between the input cell quantities (\(n_i\)) and those generated by HistoSmith (\(n_o\)).  

Finally, evaluation metrics for generative models and t-SNE embeddings were computed to assess image quality. %However, the cell quantities used to guide the generative process do not directly correspond to those in the training set.

\section{Results}
Table \ref{results:overall} presents the overall CS and CC results on CoNIC and CytoDArk. We evaluate performance using the Dice coefficient, Precision (P), Recall (R), Detection Quality (DQ), Segmentation Quality (SQ), Panoptic Quality (PQ), Coefficient of Determination (R$^2$), and Multi-class Panoptic Quality (mPQ$^+$) \cite{graham2024conic,vadori2025mind,vadori2024cisca}. The CISCA model trained on the augmented dataset $\mathcal{D} \cup \hat{\mathcal{D}}$ outperforms the one trained on the original dataset $\mathcal{D}$ across both datasets and most metrics. Specifically, it achieves an average improvement of 1.9\% across metrics on CoNIC and 3.4\% on CytoDArk0. Marked average improvements across metrics were observed for neutrophils (+5.4\%) and eosinophils (+1.4\%), as well as the hippocampus (+2.8\%) and visual cortex (+2.6\%). Notably, the visual cortex is missing from the CytoDArk0 training set, and hippocampus cells in the test set differ in appearance from those in the training set. This suggests that HistoSmith effectively enhances generalization.

\begin{figure}[t]
\centering
\begin{minipage}{0.59\textwidth} % Adjust width as needed
        \centering
	\setlength{\tabcolsep}{2pt}
	\begin{tabular}{lrrr|rrr}
	\hline
	\multicolumn{1}{c}{} & \multicolumn{3}{c|}{Test Images} & \multicolumn{3}{c}{Gen. Images} \\ \hline
	\multicolumn{1}{c}{} & \multicolumn{1}{c}{\textbf{FD}} & \multicolumn{1}{c}{\textbf{D}} & \multicolumn{1}{c|}{\textbf{C}} & \multicolumn{1}{c}{\textbf{FD}} & \multicolumn{1}{c}{\textbf{D}} & \multicolumn{1}{c}{\textbf{C}} \\
	\textbf{Colon} & 39.27 & 0.86 & 0.98 & 195.94 & 0.23 & 0.35 \\
	\textbf{Brain} & 104.05 & 0.66 & 0.98 & 316.65 & 0.41 & 0.27 \\
	- Cerebellum & 243.18 & 1.10 & 1.00 & 216.98 & 0.92 & 1.00 \\
	- Hippocampus & 243.00 & 1.52 & 1.00 & 285.19 & 0.50 & 0.90 \\
	- Aud. cortex & 83.93 & 0.99 & 1.00 & 285.42 & 0.24 & 0.33 \\ \hline
	\end{tabular}
        \captionof{table}{Assessment of the quality of generated vs. test images.}
        \label{tab:qassessment}
    \end{minipage}  
\hfill 
    \begin{minipage}{0.37\textwidth} % Adjust width as needed
        \centering
        \includegraphics[width=\textwidth]{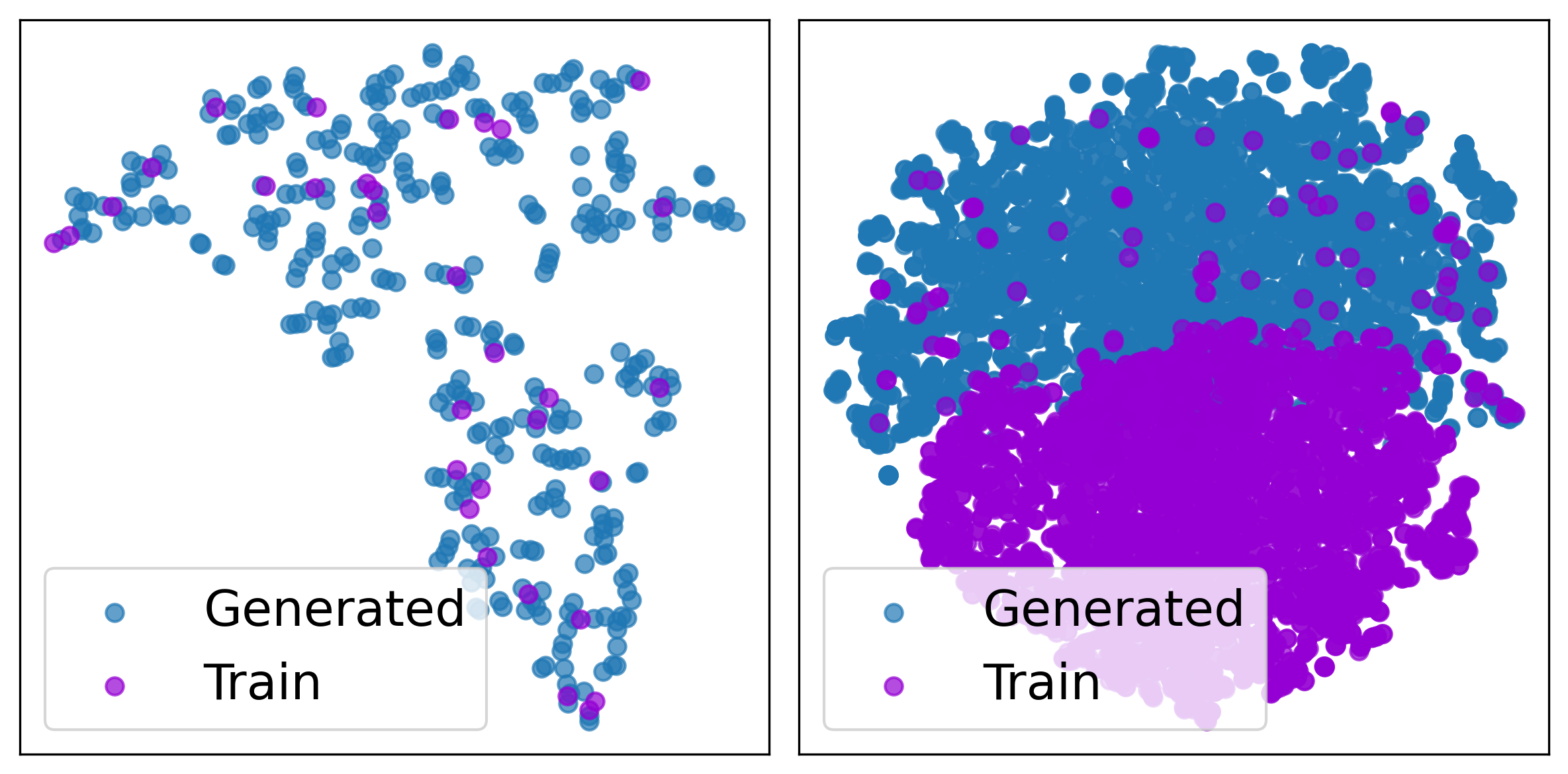} % Replace with your figure file
	\vspace{+2pt}
        \caption{t-SNE patch embeddings.}
        \label{fig:tSNEplots}
    \end{minipage}
   
\end{figure}

Fig. \ref{quantityAnalysis} presents both qualitative and quantitative analyses of the relationship between conditioning and generated cell quantities. Panels (a–c) illustrate how varying input conditioning values affects synthetic cell type semantic masks and histology images. The difference between generated (\(n_o\)) and conditioned (\(n_i\)) normalized cell counts is reported below each pair. As expected, the model effectively generates more cells as the input cell quantity increases. This trend is evident for the cerebellum (panel a) and neutrophils (panel b). However, for neutrophils, a saturation effect occurs when the cell quantity is increased excessively. In this case, we conditioned the generation process to produce H\&E-stained colon tissue with randomly sampled cell quantities for each type, as outlined in Section \ref{sec:experiments}, setting neurons/glia to zero. For neutrophils, instead of sampling, we systematically increased the cell quantity from a minimum to a maximum value.
Within the range \( \frac{1}{5} n_{\text{max}} \) to \( \frac{9}{5} n_{\text{max}} \), where \( n_{\text{max}} \) is the maximum value observed in the training set, HistoSmith generates neutrophils in proportion to the input (panel b). However, when the cell quantity is varied across the full possible range 0–1 (panel c), cells transition into the \textit{other} cell type and eventually into neurons/glia, and the image loses realism. It appears that the conditioning on tissue and cell types is almost entirely overridden to achieve the requested cell quantity, leading to the generation of non-neutrophil cell types.

Panels (d–f) provide quantitative evaluations using Bland-Altman plots, Pearson's correlation, linear regression model fits, and density distribution comparisons for 100 samples of the cerebellum, auditory cortex, and neutrophils, respectively. In the Bland-Altman plots, the y-axis represents \( n_i \) rather than the conventional average of the compared measures, as it reflects the \textquotedblleft true count\textquotedblright \ expected in the synthetic images. Overall, the bias is minimal; however, as observed qualitatively, the discrepancy between \( n_o \) and \( n_i \) increases with higher input values. Correlation is strong for the cerebellum and auditory cortex (\( r = 0.98 \)), with a linear fit closely matching the line of equality and the distribution of \( n_o \) largely aligned with that of \( n_i \), although the latter deviates from the training data distribution \( n_D \). For neutrophils (panel f, \( r = 0.39 \)), correlation is lower. While the linear trend is increasing, \( n_o \) is systematically lower than \( n_i \), as the model is consistently pushed to generate images containing neutrophils, despite their rarity in the Conic dataset (1\% of the total cells in the original training set and 2.7\% in the oversampled dataset used for training). In the distribution plots, \( n_o \) exhibits a more pronounced tail compared to \( n_i \), but it does not fully overlap. Instead, it appears stretched to the left, toward the distribution of \( n_{\mathcal{D}} \).

While Fréchet Inception Distance (FID) \cite{heusel2017gans} and Inception Score (IS) \cite{salimans2016improved} are commonly used to assess the quality of generated samples, \cite{stein2024exposing} highlights that diffusion models are unfairly penalized by the Inception network. Additionally, supervised networks do not provide a perceptual space that generalizes well for image evaluation. To address this limitation, the authors utilized a vision transformer (ViT) trained under the DINOv2 framework to extract lower-dimensional image representations for computing evaluation metrics. In Table \ref{tab:qassessment}, we report the Fréchet Distance (FD) for overall assessment, while Density (D) and Coverage (C) quantify sample fidelity and diversity, respectively \cite{naeem2020reliable}. Additionally, we visualize the t-SNE embeddings \cite{van2008visualizing} of the ViT representations for samples from the cerebellum and the colon. Generated images have a higher FD than the test set but align with state-of-the-art models \cite{stein2024exposing}. Interestingly, the cerebellum shows a lower FD in the generated images, with D and C values also well-aligned with the test data. This trend is qualitatively evident in the t-SNE plot (Fig. \ref{fig:tSNEplots}), where the generated images appear to bridge the gaps between the cerebellum training patches, which are limited to just 32. For the colon, the discrepancy in FD is the most pronounced, with D and C values significantly lower than those of the test data. However, in this study, we deliberately push the generation process beyond the high-density regions of the training and test data distributions to synthesize patches with a significantly higher presence of neutrophils and eosinophils. The limited overlap with the training patches is also evident in the t-SNE plot.

\section{Conclusions}
We introduced HistoSmith, the first LDM-based approach for histology dataset augmentation in CS and CC. Designed to address data scarcity, it also mitigates class imbalance by guiding generation toward underrepresented cell types, such as neutrophils in CoNIC. Training the CISCA model on an augmented dataset improved performance across both CoNIC and CytoDArk, with average metric gains of 1.9\% and 3.4\%, respectively. Notable improvements were seen for neutrophils (+5.4\%), eosinophils (+1.4\%), hippocampus (+2.8\%), and visual cortex (+2.6\%), indicating enhanced generalization. Additionally, HistoSmith effectively generates realistic histology images with controllable cell quantities, though accuracy depends on the conditioning range.  

These findings highlight the potential of generative data augmentation while raising key considerations. Synthetic data can improve segmentation and classification, but its effectiveness depends on the model’s generalization capacity—excessive deviation from the training distribution may introduce artifacts. Moreover, computational costs must be weighed against alternatives. While expert annotation is costly and time-intensive, motivating this study and others, it is worth exploring whether selectively annotating a well-curated subset of challenging cases offers a better trade-off between accuracy gains and cost. %Benchmarking against selective manual annotation remains essential for evaluating this trade-off.

\newpage
 \bibliographystyle{splncs04}
 \bibliography{refs}
%Author, F.: Article title. Journal \textbf{2}(5), 99--110 (2016)
%Author, F., Author, S.: Title of a proceedings paper. In: Editor,
%F., Editor, S. (eds.) CONFERENCE 2016, LNCS, vol. 9999, pp. 1--13.
%Springer, Heidelberg (2016). \doi{10.10007/1234567890}
%Author, F., Author, S., Author, T.: Book title. 2nd edn. Publisher,
%Author, A.-B.: Contribution title. In: 9th International Proceedings
%on Proceedings, pp. 1--2. Publisher, Location (2010)
%LNCS Homepage, \url{http://www.springer.com/lncs}. Last accessed 4
%\end{thebibliography}
\end{document}